\documentclass{article} 
\usepackage[nonatbib,final]{nips_2016}

\usepackage[utf8]{inputenc} 
\usepackage[T1]{fontenc}    
\usepackage{hyperref}       
\usepackage{url}            
\usepackage{booktabs}       
\usepackage{amsfonts}       
\usepackage{nicefrac}       
\usepackage{microtype}      

\usepackage{algorithm}
\usepackage{algorithmicx}
\usepackage{amsmath}
\usepackage{amssymb}
\usepackage{stmaryrd}
\usepackage{mathtools}
\usepackage[noend]{algpseudocode}
\usepackage{booktabs, siunitx}
\usepackage{multirow}
\usepackage{pgf}
\usepackage{tikz}
\usepackage{verbatim}
\usetikzlibrary{arrows,automata}

\DeclareMathOperator*{\argmax}{argmax}
\DeclarePairedDelimiter{\ceil}{\lceil}{\rceil}
\newcommand{\poly}{\mathrm{poly}}

\def\BState{\State\hskip-\ALG@thistlm}

\title{Log-time and Log-space Extreme Classification}

\author{
    Kalina Jasinska\thanks{Work done while the author was visiting Microsoft}\\
  Poznan Univeristy of Technology\\
  \texttt{kjasinska@cs.put.poznan.pl} \\
  \And{} Nikos Karampatziakis\\
  Microsoft Research\\
  \texttt{nikosk@microsoft.com} 
}

\begin{document}

\maketitle

\begin{abstract}
We present LTLS, a technique for multiclass and multilabel prediction that can perform training and inference in logarithmic time and space. LTLS embeds large classification problems into simple 
structured prediction problems and relies on efficient dynamic programming algorithms for inference.
We train LTLS with stochastic gradient descent on a number of multiclass and multilabel datasets and
show that despite its small memory footprint it is often competitive with existing approaches.
\end{abstract}

\section{Introduction}

Extreme multi-class and multi-label classification refers to problems 
where the size $C$ of the output space is extremely large.  
This problem appears in many application areas of machine learning, such as recommendation, ranking, and language modeling.
The extreme setting brings a lot of challenges, such as, inter alia, time and space complexity of training and prediction, long tail of labels, missing labels and very few training examples per label. 
In this paper we undertake the first mentioned problem -- the complexity, and propose the first, to our best knowledge, truly log-time and log-space training and prediction algorithm that can produce its top $k$ predictions in time
$O(k\log(k)\log(C))$ 
for an output space of size $C$. To do so, we adapt ideas from structured prediction to impose an efficient search 
structure on top of any multiclass and multilabel problem. This allows us to (a) characterize when we expect our technique 
to work as well as an One-Vs-All (OVA) classifier and (b) work with any model that can be trained with (online) gradient descent
including deep neural networks.

There exist several techniques for reducing training time, prediction time, and/or model size, in comparison to a OVA classifier. One technique is sparsity which can reduce model size and sometimes training and prediction times due to fewer  operations. An example of such an approach is PD-Sparse \cite{Yen_et_all_2016}, where the authors show that it is possible to get accurate sparse models in high dimensional datasets. However sparsity  is not guaranteed to reduce the model size without severely hurting model accuracy.  Another group of methods are embedding based models. Examples of such methods are SLEEC \cite{Bhatia_et_all_2015}, LEML \cite{Yu_et_all_2015}, WSABIE \cite{Weston_et_all_2011} or Rembrandt \cite{Mineiro_Karampatziakis_2015}. These techniques can
be thought of as (supervised) dimensionality reduction followed by an OVA classifier. All these approaches still remain linear in the size of the output space during training and prediction unless additional approximations are employed, such as subsampling the negative classes.

Another group consists of tree based models. Those can be further divided into decision tree based and label tree based methods. 
Those methods reduce prediction time, but not necessary lead to models with space complexity that is logarithmic in number of labels. 
For example, a multi-label decision tree based method, FastXML \cite{Prabhu_Varma_2014} builds a tree of depth logarithmic in the number of training examples. 
Label tree based method, PLT \cite{Jasinska_et_all_2016}, has a $O(\log(C))$ training time, since an update with one training instance is applied to $O(\log(C))$ models. Even though this algorithm reduces prediction time significantly, by not querying all the models, its complexity is not $O(\log(C))$.
Multi-class logarithmic time prediction is addressed by LOMtree \cite{Choromanska_Langford_2015}, but space used by this algorithm is $O(C)$.


\section{Problem Setting}

We denote with $(x, \textbf{y})$ an instance from a multi-class or multi-label training set.
Let $x$ be a feature vector, $x \in \mathbb{R}^D$, and $\textbf{y}$ a label vector of dimension $C$, $\textbf{y}_{\ell} = 1$ when $\ell$ is relevant for $x$.
In the multi-class case $\textbf{y}$ is an indicator vector, in multi-label case $\textbf{y} \in \{0, 1\}^C$.

\section{Proposed Approach}

Our idea is to embed our multiclass/multilabel problem in a structured prediction problem where the structured label $s$ is a combinatorial object. 
The model is usually specified via a compatibility function $F(x,s;w)$ which measures how well feature vector $x$ and $s$ go together for a particular setting of the parameters $w$. 
The inference problem $\hat{s}=\argmax_{s \in {\mathcal{S}}} F(x,s;w)$ requires maximization over a combinatorial set which can sometimes be solved exactly in time $O(\poly\log(|\mathcal{S}|))$ such as when $\mathcal{S}$ is the set of spanning trees or the set of matchings of a graph. 
Our criteria  for selecting a structured prediction problem are the simplicity of algorithms for finding the top 1 and top $k$ elements of $\mathcal{S}$ according to $F(x,s;w)$.

In this paper our construction is a directed acyclic graph (DAG) $G$ that contains exactly $C$ paths from a source vertex to a sink vertex.
Every edge $e$ in the graph is associated with a learnable function $h_e(\cdot;w): \mathcal X \to \mathbb{R}$. 
Every class corresponds to a path and the model predicts the class with the highest scoring path.
As usual, the score of a path is the sum of the scores of the edges in the path. 
There are many topologies we could have selected but in this paper we are exploring a minimal one because in this way we can force every learnable parameter to train 
roughly on 25\% of the data (for problems whose class prior is close to uniform); 
thus avoiding data decimation issues.

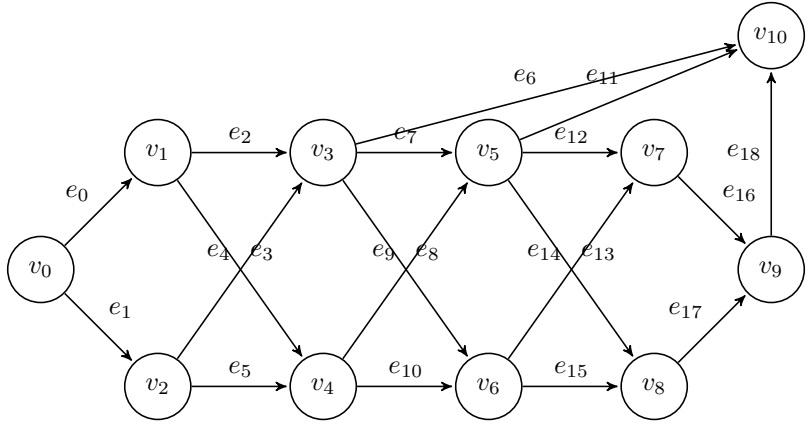
\begin{figure}[h!]
	\centering
		\begin{tikzpicture}[->,>=stealth',shorten >=1pt,auto,node distance=2.2cm,
		semithick]
		\tikzstyle{every state}=[fill=white,draw=black,text=black]
		\node[state] 		 (A)                    {$v_0$};
		\node[state]         (B) [above right of=A] {$v_1$};
		\node[state]         (C) [below right of=A] {$v_2$};
		\node[state]         (D) [right of=B] {$v_3$};
		\node[state]         (E) [right of=C] {$v_4$};
		\node[state]         (F) [right of=D] {$v_5$};
		\node[state]         (G) [right of=E] {$v_6$};
		\node[state]         (H) [right of=F] {$v_7$};
		\node[state]         (I) [right of=G] {$v_8$};
		\node[state]         (J) [below right of=H]       {$v_9$};
		\node[state]         (K) [above right of=H]       {$v_{10}$};
		\path (A) 	edge              node {$e_0$} (B)
		edge              node {$e_1$} (C)
		(B) 		edge              node {$e_2$} (D)
		edge              node {$e_3$} (E)
		(C) 		edge              node {$e_4$} (D)
		edge              node {$e_5$} (E)
		(D) edge              node {$e_6$} (K) 
		edge              node {$e_7$} (F)
		edge              node {$e_8$} (G)
		(E) edge              node {$e_9$} (F)
		edge              node {$e_{10}$} (G)		
		(F) edge              node {$e_{11}$} (K)
		edge              node {$e_{12}$} (H)
		edge              node {$e_{13}$} (I)
		(G) edge              node {$e_{14}$} (H)
		edge              node {$e_{15}$} (I)
		(H) edge              node {$e_{16}$} (J)
		(I) edge              node {$e_{17}$} (J)
		(J) edge              node {$e_{18}$} (K)
		;
		\end{tikzpicture}
	\caption{A graph $G$ for $C=22$. Source is $v_0$, sink is $v_{10}$ auxiliary is $v_{9}$. First step out of $4$ steps consists of vertices $v_1$ and $v_2$, following steps consist of next pairs of vertices.}
	\label{fig:example-graph}
\end{figure}

The graph, see an example on Figure \ref{fig:example-graph}, is a trellis of $\lfloor\log(C)\rfloor$ steps with 2 ``states'' on each step. 
The source is connected to each vertex in the first step.
An auxiliary vertex collects all vertices of the last step to one point. 
If we connect the sink vertex to the auxiliary vertex then we can only express multiclass/multilabel problems where $C$ is a power of 2.
In order to handle an arbitrary number of classes $C$ we connect the sink to one of the states at step $i$ for all $i$ such that the $i$-th (least significant) bit in binary representation of $C$ is 1. 
The upper bound of number of edges $E$ in the graph is $5\ceil{\log_2C}+1$.
The maximum scoring path can be found with dynamic programming, which in this context is known as the Viterbi algorithm.
It requires $O(E)$ steps. For each edge in order $e_1,e_2,\ldots$, the algorithm updates the highest scoring path from the source to its destination by checking  whether this edge would increase it. After all edges are processed  the algorithm backtracks to retrieve the longest path.
The top-k scoring paths can be found by a modification of the Viterbi algorithm called list Viterbi. 
We will refer to the proposed approach using name LTLS, which stands for Log-Time Log-Space.

\section{LTLS model}

A path $s$ is a vector of length $E$, with $s_e = 1$ if edge $e$ is a part of this path, and $s_e = 0$ otherwise, such that one can reach from the source vertex to the auxiliary vertex in the graph $G$ using all, and only, edges in $s$. 
We say that edge $e \in s$ if and only if $s_e = 1$.
There are exactly $C$ paths in graph $G$.
All paths in a graph $G$ stacked horizontally form a matrix $M_G$ of dimensionality $C \times E$.
Each label $\ell$ in $\mathcal{L} = (1, 2, \ldots, C)$ is exclusively assigned to a path $s(\ell_c)$. 

Given a feature vector $x$ of dimension $D$ and model weights $w$
every edge $e$ gets a value $h_{e}(w, x)$. 
Values for all the edges form a $E$-dimensional vector $h(w, x)$.
The score of the model for a label $\ell$ is the score for the corresponding path $s(\ell)$
\begin{equation}
F(x, s; w) = \sum_{e \in s}{h_{e}(w, x)} = s\cdot h(w, x),
\label{eg:F}
\end{equation}
all the label scores are based on the same $h(w, x)$.

Therefore the model is a low-rank model $\textbf{f} = M_Gh(w,x)$ embedding the label vector of dimension $C$ to a $E$-dimensional vector, such that $E$ is $O(\log(C))$. The decoding matrix $M_G$ is designed in such a way that finding the highest or $k$ highest values in \textbf{f} is fast, i.e. is of order of $E$. 

\subsection{Underlying models}

LTLS can use various learnable functions to estimate edge values $h(w, x)$.
The most basic one may use a linear model to predict each edge weight.
Then the weights become $W \in \mathbb{R}^{E \times D}$, and the low rank models gets a form  $\textbf{f} = M_G Wx$.

While we can show that if a well performing OVA linear model $V\in \mathbb{R}^{C \times D}$ can be approximated by $M_G W$ our approach will perform well (details omitted) this assumption is not always practical. Fortunately, our approach can also be used as an \emph{output layer} of a deep network, where the scores of the edges $h(w,x)$ are given by a deep network while still being able to perform inference and backpropagate through this layer in $O(\log(C))$.

\section{Optimization}

One of our goals is to have logarithmic training time. For multiclass classification
this is easy even for multinomial logistic regression because the trellis graph can 
compute the log partition function $\log\sum_{i=1}^C\exp(F(x,s(\ell_i);w))$ efficiently. Backpropagation (also known as the forward-backward algorithm in this context) can be used to compute derivatives for all parameters. This is what we use when the underlying
model is a deep network.

For multilabel classification we restrict ourselves to linear predictors and use a loss that induces (dual) sparsity. We use the \emph{separation ranking loss} \cite{Crammer_Singer_2003, Yen_et_all_2016}, being zero if all the positive labels $\mathcal{P}(\textbf{y})$ for given instance $(x, \textbf{y})$ have higher scores than all the negative labels $\mathcal{N}(\textbf{y})$ plus a margin, and being the difference between highest scoring negative label $\ell_n$ score $F(\cdot, s(\ell_n), w)$ plus a margin and lowest scoring positive label $\ell_p$ score $F(\cdot, s(\ell_p), w)$. Formally,

$$
L(w, \textbf{y}) = \max_{\ell_n \in \mathcal{N}(\textbf{y})} \max_{\ell_p \in \mathcal{P}(\textbf{y})} (1 + F(\cdot, s(\ell_n), w) - F(\cdot, s(\ell_p), w))_+.
$$

This loss also works for multiclass problems so we will use it for all our experiments
when the underlying model is linear. 
Finding the loss requires finding scores of two labels only, $\ell_p$ and $\ell_n$. 
Those can be found efficiently. 
Getting a score  $F(\cdot, s(\ell), w)$ for a given label $\ell$ is $O(E)$.
In the multiclass case $F(\cdot, s(\ell_p), w)$ is the score of the one positive class.
In multilabel case $F(\cdot, s(\ell_p), w) = \max(\{F(\cdot, s(\ell), w) : \ell \in \mathcal{P}(\textbf{y})\})$.
Since in extreme classification $|\mathcal{P}(\textbf{y})| \ll C$, this step is fast.
To find $\ell_n$ in multiclass case it is sufficient to find the 2 longest paths, in multilabel the $|\mathcal{P}(\textbf{y})| + 1$ longest paths, and determine which of them is negative. 

Since the label score $F(\cdot, s(\ell), w)$ is defined \ref{eg:F} as a sum of edge scores, we need to update only the learnable functions $h_e(\cdot;w)$ for edges $e$ in
the symmetric difference of $s(l_p)$ and $s(l_n)$. 

We use stochastic gradient descent with averaging to minimize the loss. For the linear model the gradient is 0 if the loss is 0 and otherwise it is $x$ for the models on the edges used only by the positive path and $-x$ for the models on the edges used only by the negative path.

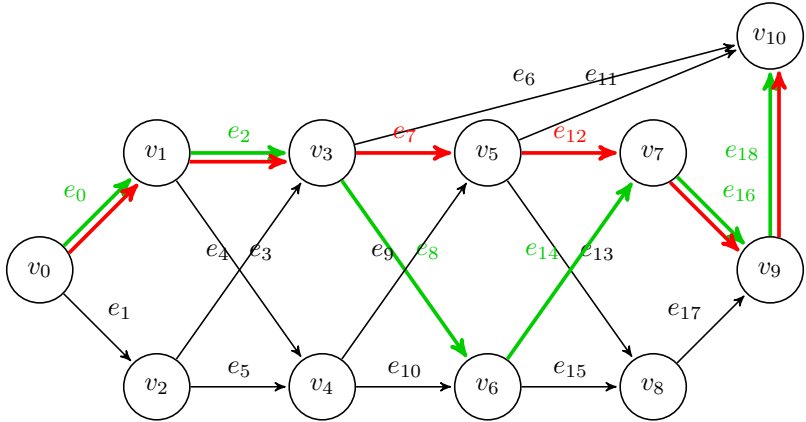
\begin{figure}[h!]
	\centering
	\begin{tikzpicture}[->,>=stealth',shorten >=1pt,auto,node distance=2.2cm,
	semithick]
	\tikzstyle{every state}=[fill=white,draw=black,text=black, scale=1]

			\node[state] 		 (A)                    {$v_0$};
			\node[state]         (B) [above right of=A] {$v_1$};
			\node[state]         (C) [below right of=A] {$v_2$};
			
			\node[state]         (D) [right of=B] {$v_3$};
			\node[state]         (E) [right of=C] {$v_4$};
			
			\node[state]         (F) [right of=D] {$v_5$};
			\node[state]         (G) [right of=E] {$v_6$};
			
			\node[state]         (H) [right of=F] {$v_7$};
			\node[state]         (I) [right of=G] {$v_8$};
			
			\node[state]         (J) [below right of=H]       {$v_9$};
			
			\node[state]         (K) [above right of=H]       {$v_{10}$};
	
	\path 	(A) edge [black!20!green,line width=1.4pt] node {$e_0$} (B)
				edge              node {$e_1$} (C)
			(A.30) edge [red,line width=1.4pt] node {} (B.240)
			(B) edge [black!20!green,line width=1.4pt] node {$e_2$} (D)
				edge              node {$e_3$} (E)
			(B.345) edge [red, line width=1.4pt] node {} (D.195)
			(C) edge              node {$e_4$} (D)
				edge              node {$e_5$} (E)
			(D) edge              node {$e_6$} (K) 
				edge [red,line width=1.4pt] node {$e_7$} (F)
				edge [black!20!green,line width=1.4pt] node {$e_8$} (G)
			(E) edge              node {$e_9$} (F)
				edge              node {$e_{10}$} (G)		
			(F) edge              node {$e_{11}$} (K)
				edge [red, line width=1.4pt] node {$e_{12}$} (H)
				edge              node {$e_{13}$} (I)
			(G) edge [black!20!green,line width=1.4pt] node {$e_{14}$} (H)
				edge              node {$e_{15}$} (I)
			(H) edge [black!20!green,line width=1.4pt] node {$e_{16}$} (J)
			(H.300) edge [red,line width=1.4pt] node {} (J.150)
			(I) edge              node {$e_{17}$} (J)

			(J) edge [black!20!green,line width=1.4pt] node {$e_{18}$} (K)
			(J.75) edge [red,line width=1.4pt] node {} (K.285)
	;
	\end{tikzpicture}
	\caption{A graph with example lowest scoring positive path (green) and highest scoring negative path (red). Learnable functions on edges $e_7$ and $e_{12}$ get a negative update, on $e_8$ and $e_{14}$ a positive, $e_0$, $e_2$, $e_{17}$ and $e_{18}$ are not updated.}
	\label{fig:graph-with-paths]}
\end{figure}

\subsection{Label-path assignment policy}
Since the decompression matrix $M_G$ structure is fixed to enable fast inference, bipartite matching between labels $\ell \in \mathcal{L}$ and paths $s \in \mathcal{S}$ becomes an important issue. 
To keep the training online and fast we could not propose a very complex method for finding a good path for each class. A simple approach is 
once an instance $(x, \textbf{y})$ with an unseen label $\ell$ is encountered, we find the top $m$ paths for $x$ and assign $\ell$ to the highest ranked free path. 
If there is no free path we assign a random path.
We restrict size of the ranking, so that $m$ is $O(log(C))$, to keep the training fast.
While this increases our memory requirements to $O(C)$ (for knowing which paths are free) this memory is not for model parameters and therefore stays constant as the input
size (or the model size in case of deep networks) increases. Training time also increases to $O(\log^2(C)\log\log(C))$ but in our experiments this makes no difference.

\section{Experiments}

This section presents an experimental evaluation of LTLS
\footnote{Code is available at \url{https://github.com/kjasinska/ltls}}. 
First we report the results of LTLS with a simple linear model on each edge and separation ranking loss.
We have run LTLS on the datasets used in \cite{Yen_et_all_2016}, where one can find a comparison of a set of multi-class and multi-label algorithms in terms of precision@1, training and prediction time, and model size. 
In the reported training times bear in mind that LTLS implementation is at the moment in Python, while other algorithms are implemented in compiled languages.

In Tables \ref{tab:multiclass} and \ref{tab:multilabel} we compare LTLS with LOMtree, FastXML, and LEML, for which we report the results from \cite{Yen_et_all_2016}. 
In case of multi-class problems on all except one dataset LTLS gets results comparable to the LOMtree, while creating a smaller model and providing the predictions faster (even though LTLS is currently in Python). 
The results using described assignment policy are significantly better than using random assignment.

A low result on the ImageNet is a good starting point for analysis when does LTLS work, and when does not.
The ImageNet dataset is quite dense in comparison to other multi-class datasets used in the experiments, average number of features active for an example is 308 out of 1000.
A model build by LTLS, weight matrix $w$, is dense -- there are nearly no zero elements.
This is because learnable function on each edge must learn many classes.
Class distribution is fairly balanced in case of this dataset, so a classifier on each edge is being updated with many examples.
Therefore we see that the underlying linear model on each edge it too simple to learn to distinguish between classes when the feature space is small and dense.

To verify the hypothesis that the poor result stems from the fact that the underlying model is too simple, we have used LTLS with a deep network. 
We have used a network with $E$ outputs to predict edge weights, and LTLS as an output layer, decoding $E$ outputs to $C$ classes. 
With a network with 2 layers, 500 hidden units in each, and ReLU nonlinearities, after 10 iterations of training we have reached $0.0507$ test precision.

On two of the multi-class datasets, LSHTC1 and Dmoz, we have observed that LTLS overfitted. We have trained by adding an $L_1$ regularization term in the objective with strength $\lambda$. This simply means predicting with soft-thresholded weights $w$:
\begin{equation*}
st(w_{ij}, \lambda) = \begin{cases}
					w_{ij} - \lambda 		& 	w_{ij} > \lambda\\
					w_{ij} + \lambda 		& 	w_{ij} < -\lambda\\
					0 						&   |w_{ij}| \leq \lambda.
				\end{cases}
\end{equation*}

For multi-label datasets the results are mixed. LTLS has performed well on rcv1-regions and LSHTCwiki, especially taking into account also prediction time and model size in case of the biggest dataset. On Eur-Lex we have encontered underfitting.

So far we have compared LTLS, a very limited model in terms of model size, with methods without such constraint. 
In table \ref{tab:naive} we present comparison of LTLS results with a naive baseline algorithm having the same model size and $O(\log(C))$ prediction time.
We have trained a 1-vs-All classifier for $E$ most frequent labels in each dataset.
As a binary classifier we have used L2-regularized Logistic Regression with tuned regularization constant.
We report an upper bound for the result, result of the naive baseline and LTLS result.

\section{Conclusions}

We have presented LTLS the first log-time and log-space technique for extreme classification. 
By embedding extreme problems into structured prediction we are able to address both time and 
space complexities while providing clear connections with low rank models and ways to incorporate
deep learning into extreme classification. Many of our design choices have been motivated purely 
from convenience and leave many interesting questions on the impact of these choices as open 
questions for future work. 

\begin{table}[h]
	
	\caption{Results on multi-class datasets. We mark with an asterisk * results where LTLS cannot fit the data, and with a dagger $^\dagger$ results of a L1-regularized LTLS model.}
	\label{tab:multiclass}
	\centering
	\begin{tabular}{lrlrrr}
		\textbf{Sector}   & \textbf{} & \textbf{}               & \textbf{LTLS} & \textbf{LOMtree} & \textbf{FastXML} \\
		\midrule
		\#examples        & 8658      & precision@1             & 0.8845        & 0.8210            & 0.8490           \\
		\#features        & 55197     & prediction time {[}s{]} & 0.14          & 0.16              & 0.25             \\
		\#classes         & 105       & model size {[}M{]}      & 5.91          & 17.00             & 7.00             \\
		\textbf{aloi.bin} & \textbf{} & \textbf{}               & \textbf{}     & \textbf{}         & \textbf{}        \\
		\midrule
		\#examples        & 100000    & precision@1             & 0.8224        & 0.8947            & 0.9550           \\
		\#features        & 636911    & prediction time {[}s{]} & 1.00          & 1.59              & 10.99            \\
		\#classes         & 1000      & model size {[}M{]}      & 102           & 106               & 992              \\
		\textbf{LSHTC1}   & \textbf{} & \textbf{}               & \textbf{}     & \textbf{}         & \textbf{}        \\
		\midrule
		\#examples        & 83805     & precision@1             & $^\dagger$0.0950       & 0.1056            & 0.2166           \\
		\#features        & 347255    & prediction time {[}s{]} & 0.65          & 6.84              & 6.33             \\
		\#classes         & 12294     & model size {[}M{]}      & 260           & 744               & 308              \\
		\textbf{ImageNet} & \textbf{} & \textbf{}               & \textbf{}     & \textbf{}         & \textbf{}        \\
		\midrule
		\#examples        & 1261404   & precision@1             & *0.0075       & 0.0537            & 0.0648           \\
		\#features        & 1000      & prediction time {[}s{]} & 15.03         & 37.70             & 139.00           \\
		\#classes         & 1000      & model size {[}M{]}      & 390           & 35                & 914              \\
		\textbf{Dmoz}     & \textbf{} & \textbf{}               & \textbf{}     & \textbf{}         & \textbf{}        \\
		\midrule
		\#examples        & 345068    & precision@1             & $^\dagger$0.2304       & 0.2127            & 0.3840           \\
		\#features        & 833484    & prediction time {[}s{]} & 5.24          & 28.00             & 57.10            \\
		\#classes         & 11947     & model size {[}M{]}      & 197           & 1800              & 1500       
	\end{tabular}
\end{table}

\begin{table}[h]

	\caption{Multi-label results. We mark with an asterisk * results where LTLS cannot fit the data.}
	\label{tab:multilabel}
	\centering
\begin{tabular}{lrlrrr}
	\textbf{Bibtex}      & \textbf{} & \textbf{}               & \textbf{LTLS} & \textbf{LEML} & \textbf{FastXML} \\
	\midrule
	\#examples           & 5991      & precision@1             & 0.2719        & 0.6401        & 0.6414           \\
	\#features           & 1837      & prediction time {[}s{]} & 0.07          & 0.22          & 0.09             \\
	\#classes            & 159       & model size {[}M{]}      & 264.0         & 8.6           & 27.0             \\
	\textbf{rcv1regions} & \textbf{} & \textbf{}               & \textbf{}     & \textbf{}     & \textbf{}        \\
	\midrule
	\#examples           & 20835     & precision@1             & 0.8964        & 0.9628        & 0.9328           \\
	\#features           & 47237     & prediction time {[}s{]} & 0.52          & 2.52          & 0.82             \\
	\#classes            & 225       & model size {[}M{]}      & 6.15          & 205.00        & 14.60            \\
	\textbf{Eur-Lex}     & \textbf{} & \textbf{}               & \textbf{}     & \textbf{}     & \textbf{}        \\
	\midrule
	\#examples           & 15643     & precision@1             & *0.0559       & 0.6782        & 0.6730           \\
	\#features           & 5000      & prediction time {[}s{]} & 0.24          & 42.24         & 1.00             \\
	\#classes            & 3956      & model size {[}M{]}      & 1.79          & 78.00         & 324.50           \\
	\textbf{LSHTCwiki}   & \textbf{} & \textbf{}               & \textbf{}     & \textbf{}     & \textbf{}        \\
	\midrule
	\#examples           & 2355436   & precision@1             & 0.2240        & 0.2846        & 0.7828           \\
	\#features           & 2085167   & prediction time {[}s{]} & 5.43          & 2896.00       & 164.80         \\
	\#classes            & 320338    & model size {[}M{]}      & 769           & 10400         & 8900       
\end{tabular}
\end{table}

\begin{table}[t]
	\caption{Naive baseline precision@1 results compared to LTLS.}
	\label{tab:naive}
	\centering
\begin{tabular}{lllll}
\multirow{2}{*}{\textbf{}} & \multicolumn{1}{c}{\multirow{2}{*}{\textbf{\begin{tabular}[c]{@{}c@{}}LTLS\\  \#edges\end{tabular}}}} & \multicolumn{2}{c}{\textbf{top-\#edges}}                              & \multicolumn{1}{c}{\multirow{2}{*}{\textbf{}}} \\
                           & \multicolumn{1}{c}{}                                                                                  & \multicolumn{1}{c}{\textbf{oracle}} & \multicolumn{1}{c}{\textbf{LR}} & \multicolumn{1}{c}{\textbf{LTLS}}                           \\
\midrule                         
\textbf{sector}  & 28    & 0.2362 & 0.2248 & \textbf{0.8945}\\
\textbf{aloi.bin}& 42    & 0.0275 & 0.0274 & \textbf{0.8224}\\
\textbf{LSHTC1}  & 56    & 0.1463 & \textbf{0.0966} & \textbf{0.0950}\\
\textbf{imageNet}& 42    & 0.0697 & \textbf{0.0340} & 0.0075\\
\textbf{Dmoz}    & 61    & 0.3507 & \textbf{0.2376} & \textbf{0.2304}\\
\midrule
\textbf{bibtex}  & 34    & 0.7126 & 0.2220 & \textbf{0.2719}\\
\textbf{rcv1-regions}& 34 & 0.8644 & 0.6576 & \textbf{0.8964}\\
\textbf{Eur-Lex}     & 52 & 0.6672 &\textbf{0.1262} & 0.0579\\
\textbf{LSHTCwiki}   & 81 & 0.2520 & 0.0314 & \textbf{0.2240}
\end{tabular}
\end{table}

\subsubsection*{Acknowledgments}

Kalina Jasinska is also supported by the Polish National Science Centre under grant no. 2013/09/D/ST6/03917. 
Some experiments were run in Poznan Supercomputing and Networking Center under computational grant no 243.

\bibliography{ltls}
\bibliographystyle{plain}

\end{document}